\documentclass[11pt]{article}

\usepackage[preprint]{acl}

\usepackage{newtxtext} 
\usepackage{newtxmath} 
\usepackage{latexsym}
\usepackage{booktabs}
\usepackage{tabularx}
\usepackage{multirow}
\usepackage{xcolor}
\usepackage{booktabs}   
\usepackage{colortbl}   

\definecolor{lawblue}{HTML}{1F77B4}   
\definecolor{laworange}{HTML}{FF7F0E} 
\definecolor{lawgreen}{HTML}{2CA02C}  
\definecolor{promptgray}{RGB}{242, 238, 230}

\definecolor{sbGreen}{HTML}{4C9A2A}   
\definecolor{sbRed}{HTML}{D65F5F}     
\definecolor{sbYellow}{HTML}{EECA5D}  
\definecolor{sbBlue}{HTML}{4878D0}    

\usepackage[T1]{fontenc}

\usepackage[utf8]{inputenc}
\usepackage{enumitem}
\setlist[itemize]{
    noitemsep, 
    leftmargin=*, 
}
\setlist[enumerate]{
    noitemsep, 
    leftmargin=*, 
}
\usepackage{microtype}

\usepackage{inconsolata}

\usepackage{graphicx}
\usepackage[dvipsnames]{xcolor}
\usepackage{xeCJK}

\usepackage{enumitem}
\setlist[itemize]{
    noitemsep, 
    leftmargin=*, 
}

\title{LexKairos: Benchmarking Legal Temporal Capabilities in LLMs}

\author{
  \textbf{Chenyang Li}\textsuperscript{1}\thanks{Research done during internship at Tsinghua University.}, 
  \textbf{Zejia Feng}\textsuperscript{2}, 
  \textbf{Yuqin Huang}\textsuperscript{3}, 
  \textbf{Yuxiao Ye}\textsuperscript{3}, 
  \textbf{Huiyuan Xie}\textsuperscript{3}\thanks{Corresponding author.} \\
  \textsuperscript{1}University of Oxford
  \textsuperscript{2}Sun Yat-Sen University
  \textsuperscript{3}Tsinghua University \\
  \texttt{chenyang.li@cs.ox.ac.uk, xieh@tsinghua.edu.cn}
}
\begin{document}
\maketitle
\begin{abstract}
Large language models (LLMs) have demonstrated strong performance across a wide range of legal tasks. In legal practice, time is a critical concept that governs the validity of statutes, the progression of legal cases, and the enforcement of procedural deadlines. However, legal temporal capabilities remain underexplored in existing legal AI benchmarks. To address this gap, we propose \textsc{LexKairos}, a comprehensive benchmark for evaluating the temporal capabilities of LLMs in the Chinese legal context across three dimensions: statutory temporal knowledge, case temporal modeling, and statute-case temporal reasoning. \textsc{LexKairos} comprises nine sub-tasks drawn from real-world Chinese judicial cases and statutes. We conduct systematic evaluations of eight LLMs under multiple inference settings, including vanilla, Chain-of-Thought (CoT), and thinking modes. Our results show that Gemini-3-Flash achieves the strongest overall performance, yet even the best-performing model exhibits notable limitations on tasks demanding precise time-sensitive statutory metadata recall or complex reasoning in time limits, indicating that legal temporal knowledge and reasoning remain open challenges for current LLMs.\footnote{Data and code are available at \url{https://github.com/thunlp/LexKairos}.}

\end{abstract}

\section{Introduction}

Large language models (LLMs) have rapidly advanced and are increasingly applied to legal tasks. Recent models have demonstrated strong performance across a wide range of legal tasks, such as legal judgment prediction (LJP) \cite{cail2018, ljp_in_english, clc-uket}, similar case retrieval~\cite{Pretraining_for_LCR} and structured legal reasoning \cite{syllogism, IRAC, lexchain}. However, existing legal AI research has primarily focused on the semantic and logical dimensions of legal materials and procedures\textemdash the ``what'' and ``how'' of legal analysis\textemdash such as identifying legal elements and modeling legal reasoning processes, while largely overlooking the temporal dimension: the ``when'' that governs the validity, enforceability, and procedural status of legal rights and obligations. In practice, legal systems are inherently time-conditioned. Statutes become effective at specific moments, procedural rules impose filing and response deadlines, and legal claims may arise, mature, or expire depending on the chronology of underlying events.
\begin{figure*}[t!]
    \centering
    \includegraphics[width=1\linewidth]{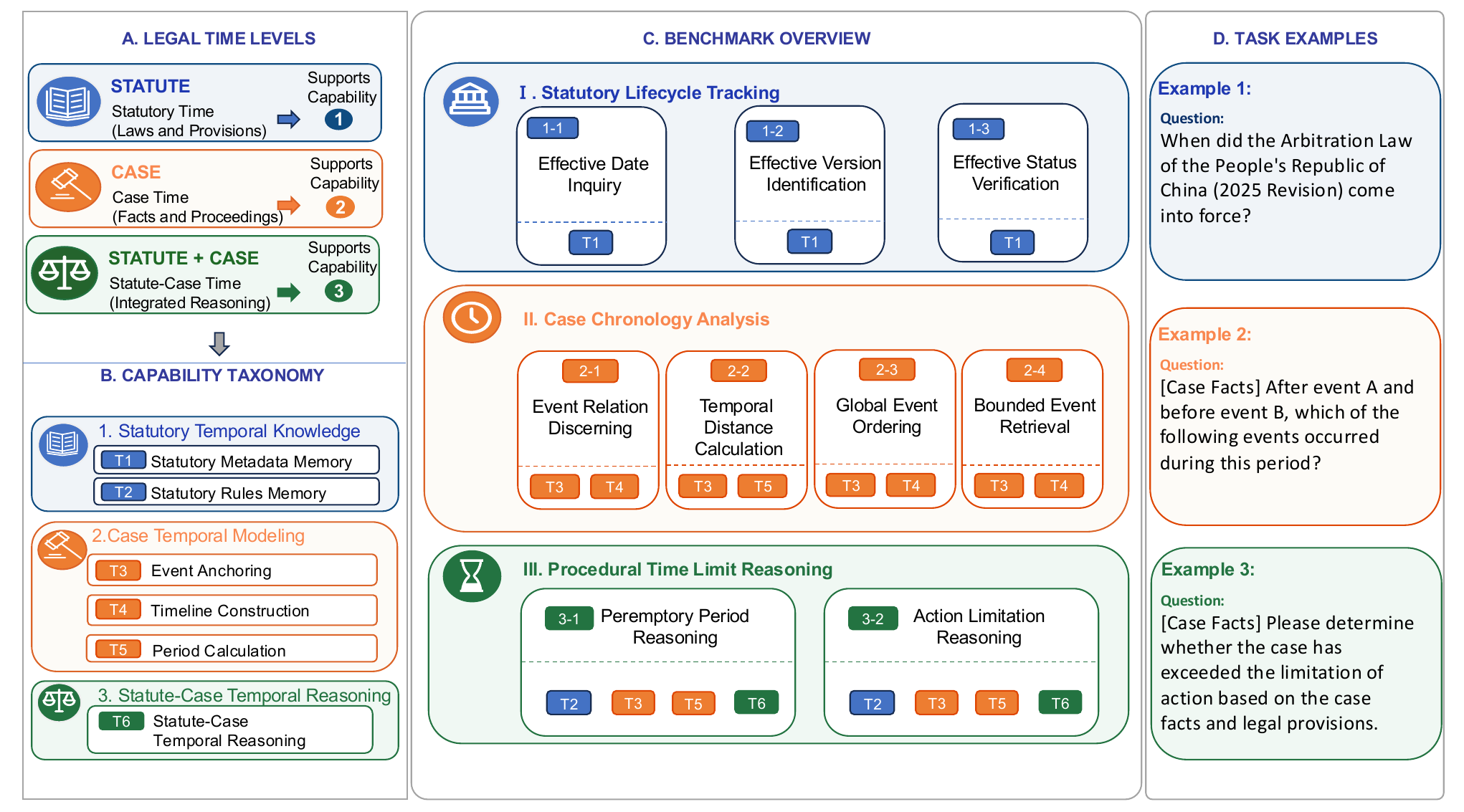}
    \caption{Task list and capability mapping in \textsc{LexKairos}. The primary tasks are mapped into nine sub-tasks, with T1--T6 representing the six specific capability aspects across three major dimensions: \textbf{\color{lawblue}{Statutory Temporal Knowledge}}, \textbf{\color{laworange}{Case Temporal Modeling}}, and \textbf{\color{lawgreen}{Statute-Case Temporal Reasoning}}.
    }
    \label{fig:architecture_overview}
\end{figure*}

To systematically characterize this temporal dimension, we conceptualize \textit{Legal Time} through two complementary levels: statutes and cases. This distinction reflects the dual role of time in legal systems: time constrains the applicability and operation of legal rules, while simultaneously governing the chronology of factual events underlying legal disputes. At the normative level, Legal Time concerns the temporal attributes of statutes, including promulgation dates, effective periods, retroactivity, and procedural deadlines imposed by legal rules. At the factual level, it governs the chronology of real-world events within legal cases, including event timestamps, event timelines, and legally significant milestones that may trigger, suspend, or extinguish legal rights and obligations.

Moreover, unlike many aspects of legal interpretation that involve substantial semantic ambiguity and discretionary judgment, legal temporal relationships are often grounded in explicit timestamps, fixed statutory periods, and formally defined procedural constraints. This relatively deterministic and verifiable structure makes Legal Time particularly well-suited for benchmarking LLMs. Temporal legal tasks require models to jointly integrate legal knowledge, temporal understanding, factual chronology, and legal reasoning under tightly constrained rule-based settings, thereby enabling precise evaluation of whether models can reason consistently about the temporal dimensions of both legal rules and legal facts.

Temporal capability has been extensively studied in general-domain settings, with benchmarks covering event ordering, commonsense duration understanding, and complex timestamp-fact relationships \cite{tempreason,TRAM}. While these efforts have advanced the evaluation of general-domain temporal reasoning, legal temporal capabilities remain largely unexplored. Existing evaluation frameworks~\cite{lawbench, lexeval} in the legal AI field primarily focus on high-level tasks such as judgment prediction and judicial analysis, in which temporal information is often overlooked or only implicitly encoded. The limited prior work that explicitly targets legal temporal capabilities has largely concentrated on isolated competencies, such as event ordering \cite{lextime} and timeline extraction \cite{lexchronos}, while overlooking statutory temporal knowledge or procedural time limit reasoning that requires joint reasoning over statutory rules and factual chronologies. These gaps highlight the need for a unified, multi-dimensional evaluation framework that captures the broad spectrum of legal temporal capabilities.
\begin{table*}[t!]
  \centering
  \small
  \begin{tabular*}{\textwidth}{@{\extracolsep{\fill}} lllcc c @{}}
    \toprule
    \textbf{Task} & \textbf{ID} & \textbf{Sub-task} & \textbf{Metric} & \textbf{Type} & \textbf{Data Volume} \\
    \midrule
    \textbf{Statutory Lifecycle } & 1-1 & Effective Date Inquiry & Accuracy & Gen & 485 \\
    \textbf{Tracking}   & 1-2 & Effective Version Identification & Accuracy & Gen & 452 \\
                                & 1-3 & Effective Status Verification & Accuracy & SLC & 480 \\
    \midrule
    \textbf{Case Chronology}    & 2-1 & Event Relation Discerning & F1-Score & SLC & 450 \\
    \textbf{Analysis}            & 2-2 & Temporal Distance Calculation & Accuracy & SLC & 550 \\
                                & 2-3 & Global Event Ordering & Accuracy & SLC & 500 \\
                                & 2-4 & Bounded Event Retrieval & Accuracy & SLC \& MLC & 594 \\
                 
    \midrule
    \textbf{Procedural Time} & 3-1 & Peremptory Period Reasoning & Accuracy & SLC & 180 \\
    \textbf{Limit Reasoning}& 3-2 & Action Limitation Reasoning & F1-Score & SLC & 109 \\
    \bottomrule
  \end{tabular*}
  \caption{Task list and data statistics in \textsc{LexKairos}. The benchmark comprises 3 primary tasks mapped into nine distinct sub-tasks. Gen, SLC, and MLC represent Generation, Single-Label Classification, and Multi-Label Classification, respectively.}
  \label{tab:data statistics}
\end{table*}

In this work, we present \textsc{LexKairos}, a comprehensive benchmark for evaluating the temporal capabilities of LLMs in the Chinese legal domain. We operationalize these legal temporal capabilities along three complementary dimensions: \textbf{Statutory Temporal Knowledge}, \textbf{Case Temporal Modeling}, and \textbf{Statute-Case Temporal Reasoning}. This benchmark suite systematically probes a model's capabilities across the dual levels of \textit{Legal Time}, statutes and cases, progressing from statute-centric knowledge recall, to fact-centric timestamp retrieval and timeline construction, and ultimately to the integrative reasoning that synthesizes both statutes and case facts.

The contributions of this work are summarized as follows:
\begin{enumerate}
\item We introduce \textsc{LexKairos}, the first dedicated benchmark for systematically evaluating legal temporal capabilities in the Chinese legal context, covering three complementary capability dimensions: statutory temporal knowledge, case temporal modeling, and statute-case temporal reasoning across nine sub-tasks.
\item We conduct systematic evaluations of eight LLMs under multiple inference settings and find that while thinking mode yields significant and consistent performance gains, existing models still exhibit notable limitations on tasks requiring precise statutory metadata recall and complex procedural reasoning.

\item A closer examination of the Effective Version Identification sub-task identifies `tag only' errors as the dominant failure mode. Further analysis of this error category reveals distinct cross-model error profiles: under vanilla inference, GPT-5.4 is characterized primarily by omission errors (59.52\% missing tag), whereas LegalOne-8B is dominated by hallucination errors (52.96\% hallucinated tag). Activating the native thinking mode substantially shifts the error distributions of both models, with each moving toward the other's predominant error profile while leaving the overall prevalence of `tag only' errors largely unchanged. Furthermore, we show that task-specific legal prompting achieves competitive performance on the Procedural Time Limit Reasoning task while producing outputs up to 70.43\% shorter, offering a highly token-efficient alternative to unconstrained thinking mode for complex legal temporal reasoning.
\end{enumerate}

\section{Related Work}

\paragraph{Temporal Capability Benchmarks} Current benchmarks have explored temporal knowledge and reasoning across multiple dimensions. For temporal knowledge, tasks querying typical event duration and common-sense temporal understanding have been widely studied \cite{timebench}. Beyond static knowledge, several works focus on world knowledge that evolves over time: TDBench \cite{tdbench} evaluates whether models can accurately retrieve facts reflecting current or historical world states, while \citet{tempreason} frames such evolving facts as ``time-event relations" and assesses models' ability to ground dynamic facts to specific timeframes. \citet{time_awareness} further proposes evaluating temporal knowledge integration via log probabilities of events across timestamps. For temporal reasoning, foundational skills such as date arithmetic have been studied \cite{timebench}, with subsequent work expanding to time zone conversion and calendar system shifts \cite{TRAM}. More complex tasks involve understanding temporal relationships between events, including chronological ordering \cite{TRAM, tempreason, timebench, time} and temporal natural language inference \cite{TRAM}. Recent benchmarks further probe reasoning robustness in multilingual contexts \cite{multilingual} and multi-hop dialogue settings \cite{time}.

\paragraph{Legal AI Benchmarks} With the rapid integration of LLMs into jurisprudence, diverse benchmarks have been developed across distinct legal dimensions. Comprehensive frameworks include LegalBench \cite{legalbench} for American law, spanning reasoning types such as issue-spotting and rule-application, and LawBench \cite{lawbench}, LexEval \cite{lexeval}, LAiW \cite{laiw}, and LexGenius \cite{lexgenius} for the Chinese legal domain, collectively covering knowledge memorization, logical inference, legal syllogism, and ethical weighing. A complementary line of research isolates granular legal skills: CAIL2018 \cite{cail2018} established foundational datasets for Legal Judgment Prediction, while more recent works probe deeper analytical tasks such as multi-step tort reasoning \cite{lexchain}, legal relation extraction \cite{lexrel}, and subparagraph-level statutory recall with historical versioning \cite{claw}.

\paragraph{Temporal Capabilities of LLMs in the Legal Domain} Recent work approaches temporal dynamics in legal AI from complementary angles. LexTempus \cite{lextempus} introduces a Dynamic Mixture of Experts architecture to mitigate temporal drift caused by evolving statutes, while LexChronos \cite{lexchronos} proposes a dual-agent system for extracting temporally ordered event timelines from unstructured judgments. Timeline generation \cite{timeline_sum} has further emerged as an integrative task to extract and organize chronological events from case communications. Beyond system development, LexTime \cite{lextime} introduces a dedicated benchmark for event ordering in the legal domain, probing LLMs' capacity to infer temporal relationships between legal events.

\section{LexKairos}

\subsection{Temporal Capability Dimensions}\label{sec: capability dimension}
The architectural design of \textsc{LexKairos} directly operationalizes the statute- and fact-driven level of Legal Time. Based on the distinction between statute-centric and fact-centric temporal demands, and the hybrid integration required when both levels interact, we partition the benchmark into three primary capability dimensions: \textbf{Statutory Temporal Knowledge} (statute-centric), \textbf{Case Temporal Modeling} (fact-centric) and \textbf{Statute-Case Temporal Reasoning} (hybrid integration).

\paragraph{Statutory Temporal Knowledge} This dimension evaluates the model's ability to recall and utilize temporal information related to legal statutes and regulations, encompassing two aspects. \textit{Statutory Metadata Memory} concerns the recall of temporal metadata associated with statutes themselves, such as effective periods and validity status. \textit{Statutory Rules Memory} concerns knowledge of temporal constraints prescribed by statutes, such as peremptory periods and statutes of limitation.

\paragraph{Case Temporal Modeling} This dimension evaluates the model's capacity to process and structure timelines within case facts, encompassing three aspects. \textit{Event Anchoring} concerns the ability to anchor legal events to specific temporal points or intervals based on the provided legal context. \textit{Timeline Construction} concerns the ability to construct a coherent chronological ordering of legal events from a given legal scenario. \textit{Period Calculation} concerns the ability to perform temporal calculations, such as determining durations between two temporal points or computing the expiration of a legal right based on its validity period.

\paragraph{Statute-Case Temporal Reasoning} This dimension evaluates the model's capacity to apply statutory temporal rules to concrete case scenarios, jointly reasoning over factual event timelines and legally prescribed time constraints to determine the temporal validity of a legal right or a legal claim.
\begin{figure}[h]
    \centering
    \includegraphics[width=1\linewidth]{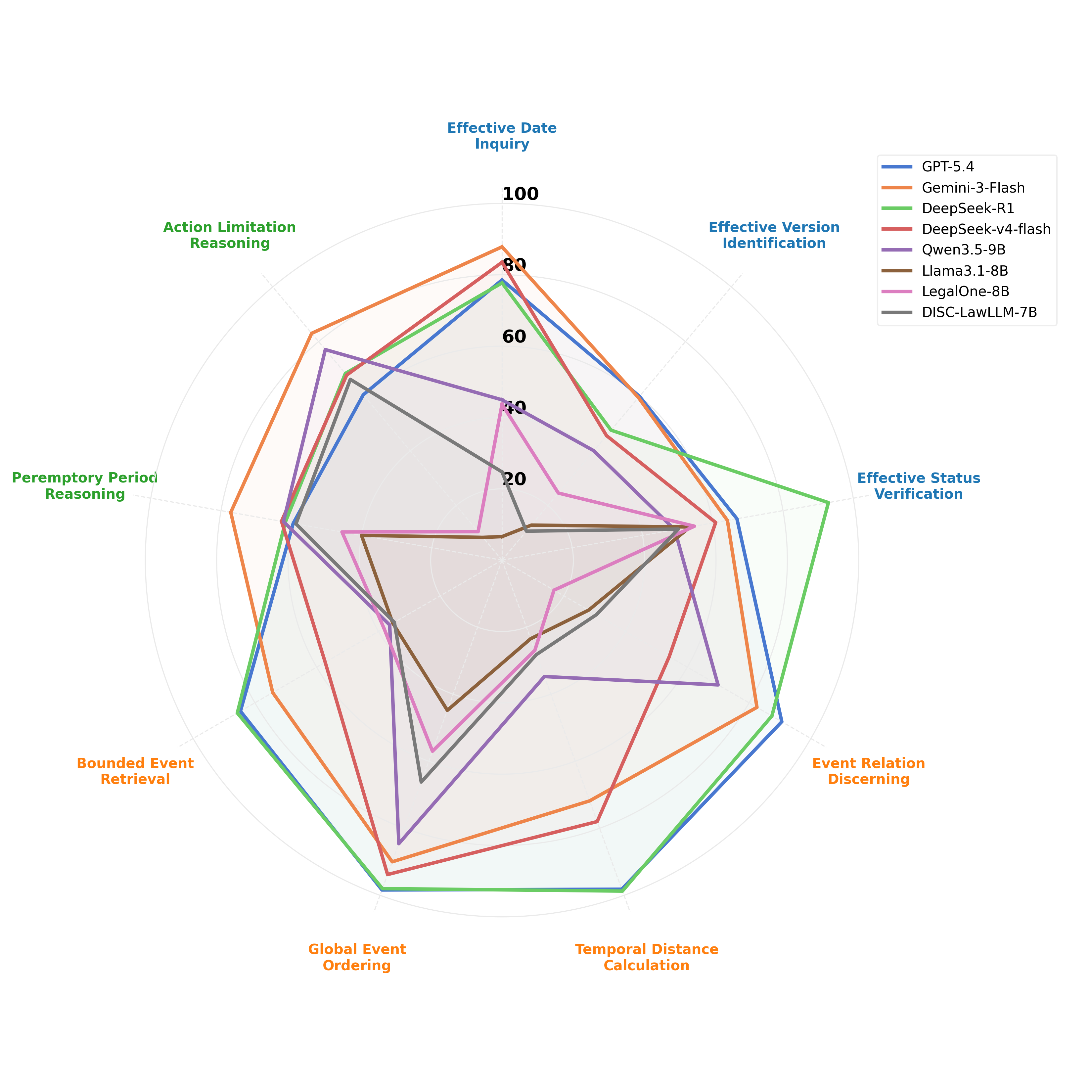}
    \caption{Results (vanilla setting) of LLMs evaluated on three tasks: 
\textcolor{lawblue}{Statutory Lifecycle Tracking}, 
\textcolor{laworange}{Case Chronology Analysis}, and 
\textcolor{lawgreen}{Procedural Time Limit Reasoning} (comprising nine sub-tasks).}
    \label{fig:radar}
\end{figure}

\subsection{Task Construction}

In alignment with the three capability dimensions outlined in Section \ref{sec: capability dimension}, each task targets one or more dimensions: Task 1 evaluates Statutory Temporal Knowledge; Task 2 evaluates Case Temporal Modeling; and Task 3 evaluates Statute-Case Temporal Reasoning, requiring models to jointly integrate the preceding two capability dimensions. Together, these three tasks consist of nine sub-tasks designed to comprehensively evaluate LLMs' legal temporal capabilities. Figure~\ref{fig:architecture_overview} provides an overview of the tasks in \textsc{LexKairos} and their corresponding capability dimensions.

\subsubsection{Statutory Lifecycle Tracking}
This task evaluates the models' Statutory Metadata Memory, focusing on the statute level of Legal Time. In judicial practice, the frequent revision, amendment, and repeal of regulations constantly alter these underlying temporal metadata. Applying a statute that is not yet in force or has already been superseded may lead to an incorrect legal basis for adjudication.
 
To assess whether models can accurately track statutory lifecycles and avoid such temporal misalignment, we design three sub-tasks:

\paragraph{Effective Date Inquiry (1-1)} Given a specific version of a statute, generate its exact effective date.

\paragraph{Effective Version Identification (1-2)} Given the name of a statute and a specific timestamp, identify the version that was in force at that time.

\paragraph{Effective Status Verification (1-3)} Given a statutory version and a specific timestamp, determine whether that version was in force at the specified time.

\subsubsection{Case Chronology Analysis}

This task evaluates the models' temporal modelling capability to perform event anchoring, timeline construction, and  period calculation. While judicial judgments are generally well structured and chronologically organized, the concrete temporal clues within factual descriptions remain densely interwoven with narrative reasoning. Missing, misordering, or miscalculating these factual milestones can fundamentally distort the legal causality and the chain of evidence required for accurate adjudication. To measure how effectively models can transform these semi-structured legal narratives into clean topological timeline, we design four sub-tasks:

\paragraph{Event Relation Discerning (2-1)} Given a case and two factual events extracted from the case, discern their temporal connection based on three standard temporal relations defined by Allen's interval relations~\cite{allens_interval}: after, meet, and during.

\paragraph{Temporal Distance Calculation (2-2)} Given a case and two factual events extracted from the case, compute the exact number of calendar days elapsed between them.

\paragraph{Global Event Ordering (2-3)} Given a case and a sequence of four legal events in the case, synthesize the available temporal clues to reconstruct their correct chronological order.

\paragraph{Bounded Event Retrieval (2-4)} Given a case and one or two extracted anchoring events, retrieve either the single temporally nearest event or all events occurring within the closed interval defined by the anchor events.

\subsubsection{Procedural Time Limit Reasoning}

This task evaluates models' Statute–Case Temporal Reasoning capability, requiring them to apply statutory temporal rules (e.g., procedural time limits) to concrete case scenarios through joint reasoning over statutory provisions and factual timelines. To evaluate this capability, we introduce the task of Procedural Time Limit Reasoning. In judicial practice, determining whether a legal right or claim falls within the applicable time limit is a fundamental threshold issue. Incorrectly applying these temporal constraints may lead to erroneous legal conclusions, such as recognizing extinguished rights or allowing time-barred claims to proceed. To assess models' ability to perform this form of legal temporal reasoning, we design two reasoning sub-tasks:

\paragraph{Peremptory Period Reasoning (3-1)} Given a case and a specific legal right, reason over absolute legal deadlines (Peremptory Periods) to determine whether the legal right has been permanently extinguished.

\paragraph{Action Limitation Reasoning (3-2)} Given a case, determine whether the statutory time limit for a plaintiff to file a valid lawsuit (Limitation of Action) has expired.

\subsection{Data Curation}

\textsc{LexKairos} is constructed from real-world Chinese judicial cases and statutes. To accommodate the varying characteristics of the tasks, we adopt different curation strategies. Task 1 instances are automatically synthesized from statutory metadata using a rule-based pipeline with manual quality filtering. Task 2 instances are built from verified civil cases through LLM-assisted event extraction and SQL-based QA generation. Task 3 instances are curated through a multi-stage hybrid pipeline combining LLM-based information extraction, rule-based filtering, and expert legal review. Dataset statistics are summarized in Table \ref{tab:data statistics}, and full details of the curation pipeline are provided in Appendix \ref{sec:data curation}.

\begin{table*}[t!]
    \centering
    \resizebox{1\linewidth}{!}{
    \setlength{\tabcolsep}{5.5pt}
    \begin{tabular}{lc cccc cccc cc}
    \toprule
    \multirow{2}{*}[-0.7ex]{\textbf{Model}} & \multirow{2}{*}[-0.7ex]{\textbf{Overall}} & \multicolumn{3}{c}{\textbf{Task 1}} & \multicolumn{4}{c}{\textbf{Task 2}} & \multicolumn{2}{c}{\textbf{Task 3}} \\
    \cmidrule(lr){3-5} \cmidrule(lr){6-9} \cmidrule(lr){10-11}
    & & \textbf{1-1} & \textbf{1-2} & \textbf{1-3} & \textbf{2-1} & \textbf{2-2} & \textbf{2-3} & \textbf{2-4} & \textbf{3-1} & \textbf{3-2} \\
    \midrule
    \textbf{GPT-5.4}            & 77.47 & 78.56 & 59.96 & 66.88 & 90.58 & 98.18 & \textbf{98.40} & 84.68 & 59.44 & 60.52 \\
    \quad w/ Thinking           & 82.75 & 80.62 & 58.63 & 90.83 & \textbf{92.80} & 98.00 & 98.20 & 86.20 & 69.44 & 70.02 \\
    \midrule
    \textbf{Gemini-3-Flash}     & 76.71 & \textbf{87.84} & 59.51 & 64.17 & 82.53 & 71.82 & 90.00 & 74.24 & \textbf{77.22} & \textbf{83.02} \\
    \quad w/ Thinking           & \textbf{84.57} & 85.77 & \textbf{60.62} & \textbf{92.92} & 88.15 & 98.55 & 98.20 & \textbf{88.72} & 74.44 & 73.78 \\
    \midrule
    \textbf{DeepSeek-R1}        & 79.78 & 77.73 & 47.57 & \textbf{92.92} & 87.37 & \textbf{98.73} & 98.00 & 85.69 & 61.67 & 68.33 \\
    \midrule
    \textbf{DeepSeek-V4-Flash}  & 67.07 & 83.51 & 45.58 & 60.83 & 54.13 & 78.00 & 93.80 & 57.24 & 62.78 & 67.74 \\
    \quad w/ Thinking           & 79.94 & 85.36 & 57.74 & 90.83 & 74.14 & 98.18 & 97.40 & 86.87 & 64.44 & 64.54 \\
    \midrule
    \textbf{Qwen3.5-9B}         & 55.45 & 44.95 & 40.04 & 49.17 & 69.94 & 34.73 & 84.60 & 36.36 & 62.22 & 77.07 \\
    \quad w/ Thinking           & 69.79 & 58.95 & 37.39 & 78.85 & 87.67 & 77.64 & 91.80 & 69.19 & 57.22 & 69.43 \\
    \midrule
    \textbf{Llama3.1-8B}        & 28.11 &  6.60 & 12.83 & 53.54 & 28.04 & 23.45 & 44.80 & 35.35 & 40.00 &  8.34 \\
    \quad w/ CoT                & 31.55 &  5.57 &  4.20 & 45.21 & 29.38 & 30.18 & 59.20 & 34.01 & 47.22 & 28.98 \\
    \midrule
    \textbf{LegalOne-8B}        & 35.33 & 43.71 & 24.56 & 54.79 & 16.82 & 26.91 & 57.00 & 38.21 & 45.56 & 10.43 \\
    \quad w/ Thinking           & 73.00 & 74.23 & 47.12 & 86.88 & 87.57 & 80.55 & 56.40 & 72.39 & 76.11 & 75.76 \\
    \midrule
    \textbf{DISC-LawLLM-7B}     & 41.13 & 24.74 & 10.62 & 50.21 & 30.57 & 28.18 & 66.20 & 34.85 & 58.66 & 66.16 \\
    \quad w/ CoT                & 38.17 & 13.61 & 14.16 & 45.00 & 31.63 & 44.36 & 63.60 & 35.02 & 49.72 & 46.44 \\
    \bottomrule
    \end{tabular}
    }
    \caption{Evaluation results on the \textsc{LexKairos} benchmark across nine distinct sub-tasks. Entries labeled with the model name alone represent the vanilla inference setting. The variant ``w/ CoT'' or ``w/ Thinking'' represent Chain-of-Thought and thinking mode enabled, respectively.}
    \label{tab:eval_results}
\end{table*}

\section{Experiments and Results}

\subsection{Baselines}
To systematically study the temporal capacities of various LLMs, we structure our benchmarks around two overarching inference paradigms based on whether the reasoning process is natively or externally triggered:

\paragraph{Vanilla Mode:} Under this baseline setting, models are expected to directly output the final answer without extra reasoning steps. For conventional LLMs, this corresponds to standard direct prompting (see Appendix \ref{sec:evaluation prompts} for detailed instructions).

\paragraph{Thinking Mode:} This setting applies exclusively to models with built-in reasoning mode. We use the identical task prompt from the Vanilla setting, but enable their intrinsic reasoning capacities via model-level configurations to let them automatically generate their inner thoughts before answering.

\paragraph{Chain-of-Thought Prompting:} This setting applies to conventional LLMs that lack a built-in thinking mode. We append a generic ``\textit{Let's think step by step}'' instruction to the standard task prompt to elicit an explicit chain of reasoning before generating the final answer.

\subsection{Models and Experimental Setup}

\paragraph{Models}
We evaluate a range of state-of-the-art models, including both open-source and proprietary models. These include GPT-5.4~\cite{gpt5_4}, DeepSeek-R1~\cite{deepseek_r1}, DeepSeek-V4~\cite{deepseekv4}, Gemini-3-Flash~\cite{google2025gemini}, Qwen3.5-9B~\cite{qwen35}, Llama-3.1-8B~\cite{llama3}. We also evaluate LegalOne-8B~\cite{legalone} and Disc-LawLLM-7B \cite{disclaw}, both of which are specifically trained on legal data. 

\paragraph{Experimental Setup}

To avoid output truncation on tasks requiring extensive contextual understanding and multi-step reasoning, we uniformly set the maximum output length to 8,000 tokens across all experimental configurations. The inference temperature is fixed at 0 to ensure deterministic decoding. For models with a built-in thinking mode, we set the thinking effort to high when evaluating their Thinking Mode performance.

\subsection{Evaluation Results}

The evaluation results on \textsc{LexKairos} are reported in Table \ref{tab:eval_results}. Overall, closed-source models generally outperform smaller-scale open-source models across most task categories and inference settings, with Gemini-3-Flash (Thinking) and GPT-5.4 (Thinking) achieving the highest overall scores of 84.57 and 82.75, respectively. Among open-source models, performance varies considerably across different model families. Disc-LawLLM-7B and LegalOne-8B, despite domain-specific fine-tuning, achieve overall scores of only 41.13 and 35.33 in the vanilla zero-shot setting, respectively, remaining well below the performance of the leading closed-source models.

Performance varies substantially across the three task categories, suggesting uneven model proficiency across different legal temporal capabilities. Across nearly all evaluated models, the sub-tasks in Case Chronology Analysis (T2) consistently achieve higher performance than those in the other two task categories. This trend suggests that current LLMs are relatively more proficient at modeling temporal information embedded in case facts, including event anchoring, timeline construction, and period calculation, than at recalling statutory temporal knowledge (Statutory Lifecycle Tracking, T1) or applying statutory temporal rules to concrete case scenarios (Procedural Time Limit Reasoning, T3).

We further investigate the impact of explicit reasoning on model performance. Enabling built-in thinking mode yields consistent performance gains overall, with larger improvements generally observed for models with weaker vanilla performance.  LegalOne-8B achieves the largest relative gain (+106.61\%), followed by Qwen3.5-9B (+25.86\%), whereas GPT-5.4 and Gemini-3-Flash exhibit comparatively modest improvements (+6.82\% and +10.25\%, respectively). This effect is particularly evident in legal temporal knowledge-intensive tasks (T1 and T3). For example, under the vanilla setting, LegalOne-8B substantially underperforms the comparable-scale general-purpose baseline Qwen3.5-9B (35.33 vs. 55.45). However, under thinking mode, LegalOne-8B surges to 73.00, outperforming Qwen3.5-9B (69.79).

In contrast, externally prompted CoT reasoning produces mixed effects across models. For example, Llama3.1-8B exhibits only modest improvement of 12.24\%, whereas Disc-LawLLM experiences performance degradation under explicit step-by-step prompting, with its overall score declining from 41.13 to 38.17.

\subsection{Analysis}

\begin{figure}[t!]
    \centering
    \includegraphics[width=1\linewidth]{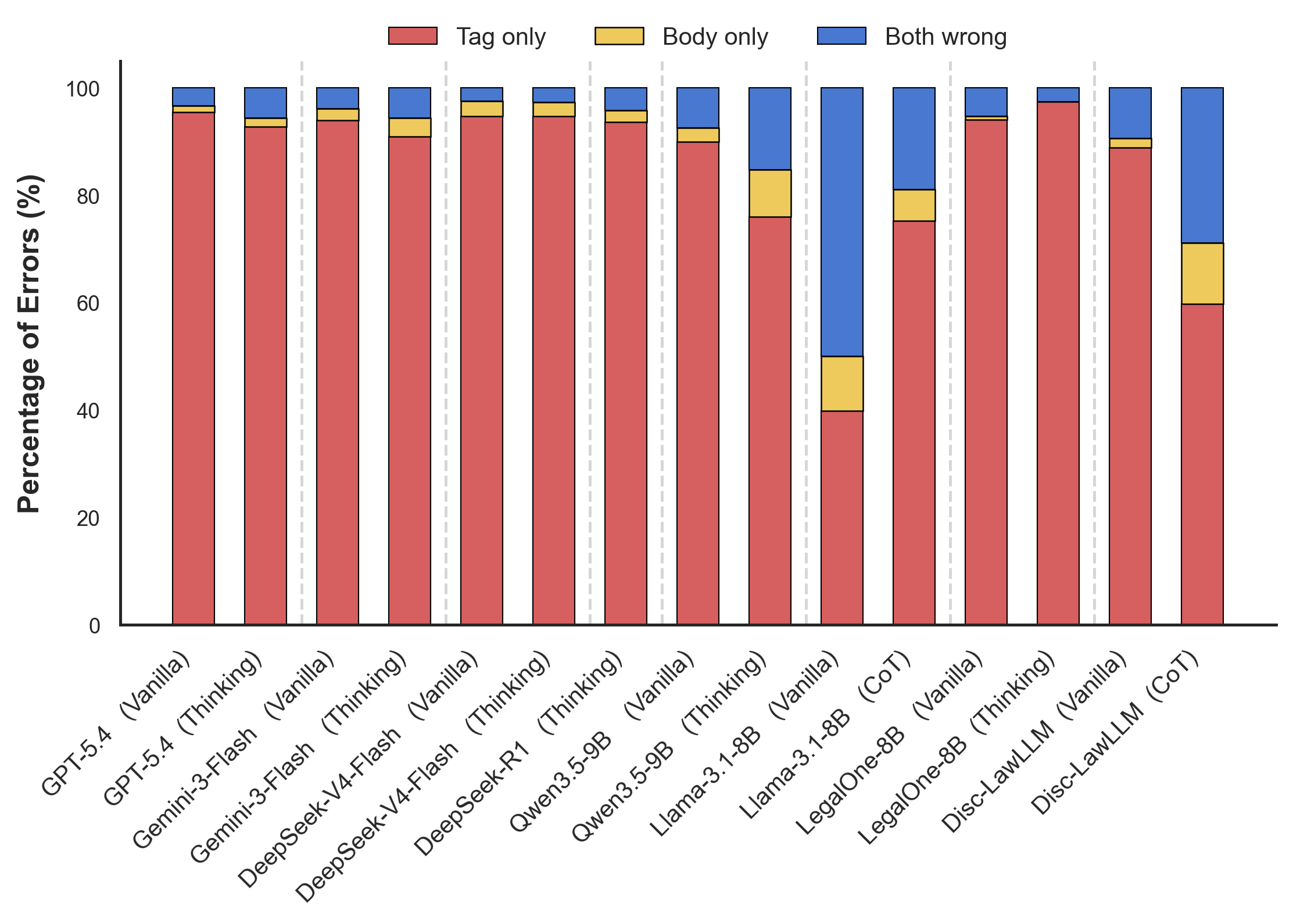}
    \caption{Error decomposition for the Effective Version Identification task (Task 1-2). 
Each stacked bar represents a model-setting pair, with error counts normalized to sum to $100\%$. 
The colored segments denote the three failure modes 
(\textcolor{sbRed}{tag only}, 
 \textcolor{sbYellow}{body only}, and \textcolor{sbBlue}{both wrong}) .
Dashed vertical lines separate distinct base models.}
    \label{fig: level2_decompose}
\end{figure}

\paragraph{Error Type Analysis on Sub-Task 1-2}
To further investigate the sources of error in the Effective Version Identification task (sub-task 1-2), we decompose model errors into three mutually exclusive categories: (1) `body only', where the model fails to produce the correct statutory title but provides a matching version tag; (2) `tag only', where the statutory title is correctly identified but the version identifier is imprecise; and (3) `both wrong', where neither component is correct. As shown in Figure \ref{fig: level2_decompose}, `tag only' errors constitute the overwhelming majority of failures across all models and settings (85.20\% on average). This finding indicates that the primary challenge of sub-task 1-2 lies not in identifying the target statute itself, but in accurately recalling its corresponding version identifier, highlighting version identifier recall as the principal bottleneck of Statutory Metadata Memory. 

\begin{figure}[t!]
    \centering
    \includegraphics[width=0.8\linewidth]{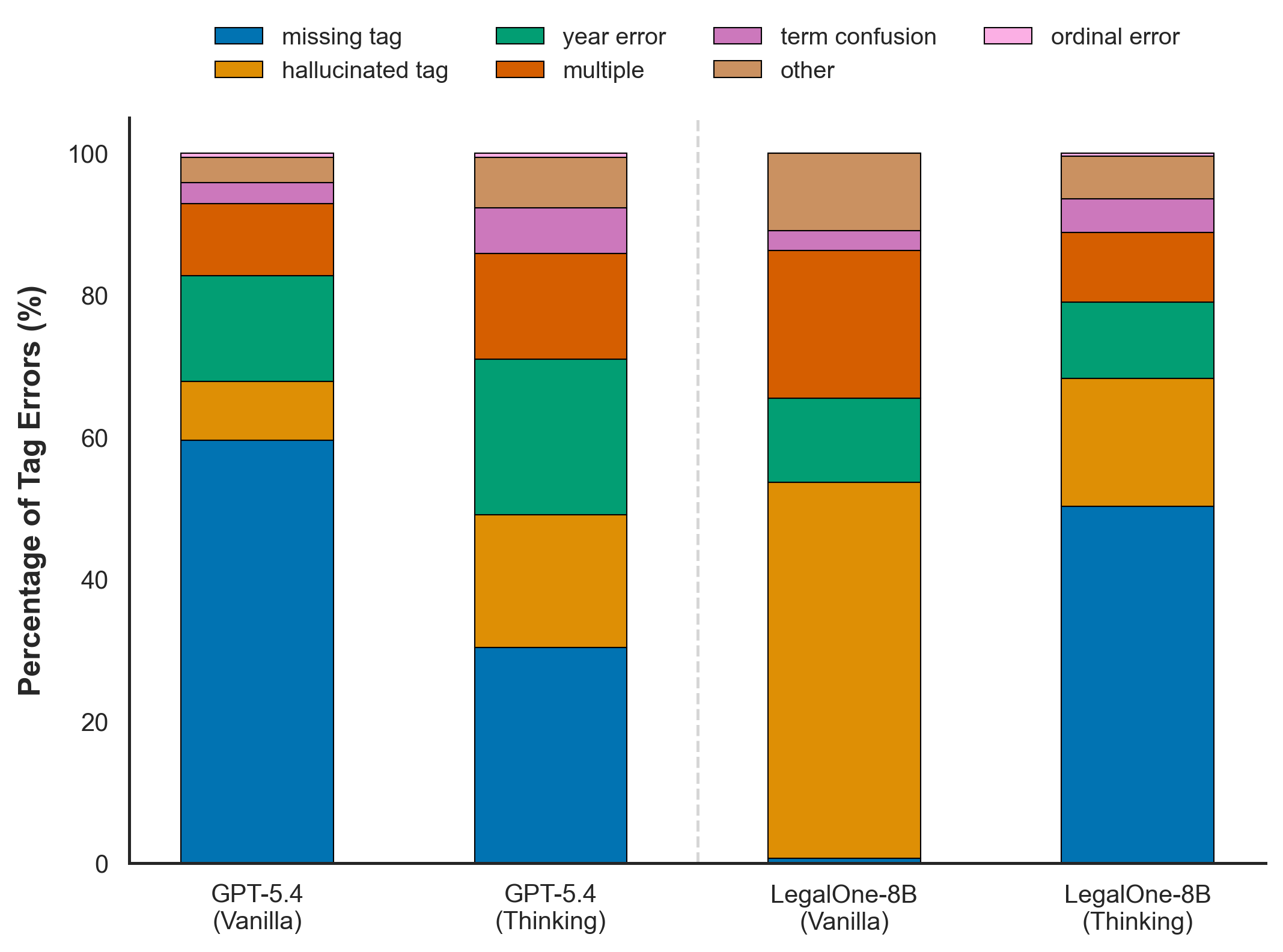}
    \caption{Sub-category distribution within `tag only' errors for different model configurations. 
Each stacked bar is normalized to $100\%$ of the total `tag only' failures for the corresponding model configuration. Dashed vertical lines distinguish separate model families.}
    \label{fig: tag only distribution}
\end{figure}

To further characterize the nature of `tag only' failures, we conduct a fine-grained error type analysis on GPT-5.4 and LegalOne-8B under both vanilla and thinking settings, classifying each `tag only' error into one of seven sub-categories: `missing tag' (version identifier absent from output), `hallucinated tag' (version identifier fabricated when none exists in the gold answer), `year error' (incorrect amendment or revision year), `term confusion' (e.g., amendment  vs. revision), `ordinal error' (incorrect amendment or revision number), `multiple' (co-occurring errors), and `other'. As shown in Figure \ref{fig: tag only distribution}, the results reveal a significant cross-model asymmetry in failure modes.

Under vanilla inference, GPT-5.4 and LegalOne-8B exhibit diametrically opposed error profiles. GPT-5.4 is dominated by `missing tag' errors (59.52\%), reflecting a conservative generation strategy in which version identifiers are frequently omitted. LegalOne-8B, by contrast, is dominated by `hallucinated tag' errors (52.96\%), reflecting a more aggressive output pattern characterized by the generation of unsupported version identifiers. This contrast reveals a pronounced cross-model asymmetry in uncertainty handling, with GPT-5.4 favoring omission and LegalOne-8B favoring fabrication when recalling statutory version metadata. 

Crucially, enabling thinking mode pushes both models toward the opposite error profile. For GPT-5.4, `missing tag' errors decrease from 59.52\% to 30.32\%, while `hallucinated tag' errors nearly double from 8.33\% to 18.70\%, indicating that extended reasoning encourages the model to attempt version generation but introduces new fabrication errors in the process. For LegalOne-8B, the pattern reverses: `hallucinated tag' errors drop sharply from 52.96\% to 17.60\%, while `missing tag' errors surge from 0.62\% to 50.21\%, suggesting that thinking mode induces greater epistemic caution, suppressing confident but incorrect generation. In both cases, thinking mode shifts the error distribution without resolving the underlying precision deficit, consistent with our earlier observation that `tag only' error rates remain persistently high regardless of inference strategy.

Finally, Qwen3.5-9B exhibits an anomalous degradation under thinking mode: overall accuracy on sub-task 1-2 decreases from 40.04\% to 37.39\%, while the proportion of `both wrong' errors rises from 7.38\% to 15.19\%. Manual inspection reveals that 58 out of 452 responses (12.83\%) are truncated before reaching the final answer block. This budget-bound bottleneck highlights a practical trade-off of unconstrained test-time compute: extended reasoning chains may exhaust the maximum token budget before producing a valid answer, an effect particularly pronounced in models whose reasoning traces expand substantially under thinking mode.

\paragraph{Truncation and Efficiency Analysis on Qwen}

\begin{figure}[t!]
    \centering
    \includegraphics[width=1\linewidth]{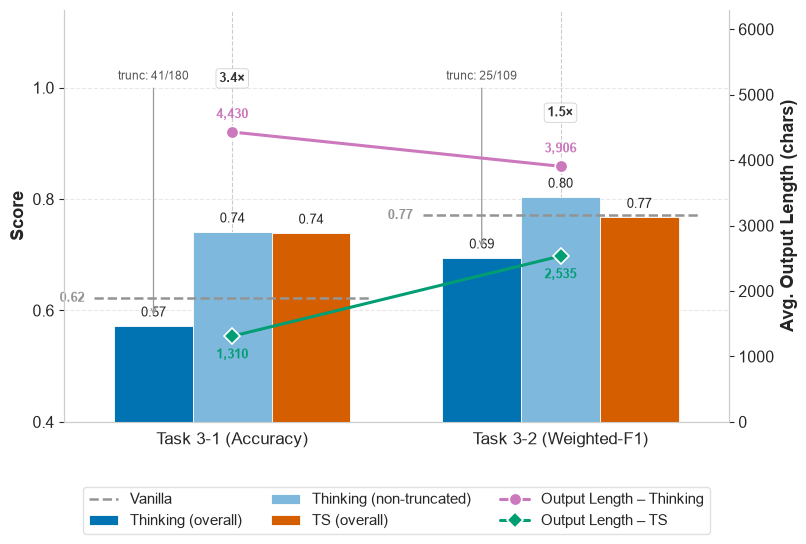}
    \caption{Performance and output length on Task 3. Bars (left axis) show scores for Thinking (overall), Thinking (non-truncated), and TS (task-specific prompting, overall), with the Vanilla baseline marked by a horizontal dashed gray line. Numbers above Thinking (overall) bars give the truncated-response fraction (trunc: n/N). Lines (right axis) show average output length in characters for Thinking and TS on sub-tasks 3-1 and 3-2; the annotated ratio indicates how much longer Thinking's average output is relative to TS.}
    \label{fig:analysis_on_qwen}
\end{figure}

We further investigate the impact of output truncation on Qwen3.5-9B under thinking mode in Task 3. As shown in Figure \ref{fig:analysis_on_qwen}, on sub-task 3-1, 41 out of 180 responses (22.78\%) are truncated before producing a valid answer, yielding zero accuracy on truncated samples and dragging the overall accuracy from 74.10\% (non-truncated) down to 57.22\%. A similar pattern holds for sub-task 3-2, where 25 out of 109 responses (22.94\%) are truncated, reducing the overall weighted-F1 from 80.34\% to 69.43\%.
Motivated by this observation, we investigate whether structured legal guidance can provide a more inference-efficient alternative to unconstrained CoT reasoning, we design task-specific legal prompts (TS) for Task 3 in consultation with legal domain experts (see Table~\ref{tab:task_3_1_ts} and Table~\ref{tab:task_3_2_ts} for detailed instructions). Applying TS leads to more concise outputs than thinking-mode, averaging 1,310 and 2,535 characters on sub-tasks 3-1 and 3-2 respectively, compared to 4,430 and 3,906 characters under thinking mode, corresponding to 70.43\% and 35.10\% reductions.

These results suggest that injecting structured legal reasoning directly into the prompt guides the model through the necessary inferential steps within a substantially more token-efficient budget, achieving performance on par with unconstrained CoT reasoning while avoiding the truncation failures that undermine thinking mode in practice.

\section{Conclusion}
In this work, we present \textsc{LexKairos}, a comprehensive benchmark for evaluating temporal capabilities in legal domain across multiple capability dimensions. We conduct a systematic study on LLMs' ability to recall, model, and reason over temporal information under legal context, and find that existing models exhibit notable limitations in recalling statutory temporal knowledge and reasoning over procedural time limits. Although thinking mode yields consistent gains across models, our fine-grained error analysis shows that it fails to resolve the underlying precision deficit in legal version identifier recall. These results highlight the need for stronger temporally aware reasoning capabilities in LLMs for legal applications.
\section*{Limitations}

A potential limitation of our work is the relatively modest size of the evaluation set for Procedural Time Limit Reasoning, which contains 180 and 109 gold-standard instances for peremptory periods and action limitations, respectively. This limited scale reflects the inherent difficulty of constructing high-quality evaluation data for complex legal temporal reasoning. Verifying procedural time limits requires legal experts to manually examine intricate interactions between statutory rules and case-specific temporal facts, making annotation highly labor-intensive and time-consuming. We therefore prioritized annotation quality and legal precision over dataset scale by relying on expert manual verification for every test instance. Future work could expand this portion of the benchmark through larger-scale expert annotation while preserving the same standard of legal rigor.

Additionally, the evaluation of Global Event Ordering is currently limited to day-level granularities. Although our structured database preserves temporal metadata at heterogeneous granularities, ranging from months to sub-day timestamps, we deliberately restrict evaluation instances to events with day-level precision. This design choice avoids chronological ambiguities that cannot be deterministically resolved from the source data. For example, ordering a month-level event (e.g., ``2012-02'') relative to a day-level event (e.g., ``2012-02-05''), or sequencing multiple events occurring on the same day without explicit timestamps, is inherently under-specified. By enforcing a minimum one-day interval between target events, we ensure that every evaluation instance admits a unique, verifiable ordering. Consequently, \textsc{LexKairos} does not yet evaluate multi-scale temporal reasoning, such as reasoning across mixed temporal granularities or within sub-day timelines. Extending the benchmark to support fine-grained and heterogeneous temporal reasoning constitutes an important direction for future work.

\bibliography{acl_latex}

\appendix

\section{Details of Data Curation}

\label{sec:data curation}
We collected the corresponding statutory temporal metadata from publicly available sources, capturing their precise effective dates, promulgation dates, and all historical revisions. A rule-based mechanism was then implemented to automatically synthesize the question-answer pairs for each sub-task. Following a manual check alongside the removal of duplicates and low-quality instances, this pipeline yielded over 400 high-quality instances for each sub-task.

All cases used in this study are sourced from publicly available court decisions published on~\citet{cjo2013}. 
 For Case Chronology Analysis in Task 2, we randomly sampled 500 civil source cases and utilized DeepSeek-V3.2  to extract explicit events, timestamps, and temporal clues (e.g., continuity and offsets) from the factual descriptions. 
For events lacking explicit timestamps, we computed their temporal anchors via extracted chronological dependencies, while integrating implicit events inferred from the overarching legal context. Following a meticulous manual audit of these temporal attributes, we constructed a structured database comprising 293 verified instances. Crucially, a single legal case typically contains multiple distinct chronological milestones and interwoven event relationships. To comprehensively evaluate LLMs' granular competencies, we executed SQL queries with diverse temporal constraints over this structured database, thereby systematically deriving multiple distinct question-answer pairs (e.g., event relation discerning, distance calculation) from a individual case.

To build the gold-standard evaluation dataset for Task 3, we implemented a unified, multi-stage curation framework across its sub-tasks. The framework initiated with coarse-grained sampling from distinct task-specific source pools: 4,000 civil cases for peremptory periods and a broader pool of 22,752 cases for action limitations. Next, a hybrid filtration phase was applied to refine these pools: for peremptory periods, structural clues were extracted via DeepSeek-V3.2 followed by a rule-based filter to retain 260 refined candidates; for action limitations, the pool was first compressed into 109 core instances via keyword intersection and similarity-based de-duplication, and then enriched with DeepSeek-V3.2 extraction. Finally, these refined candidate cases were consolidated and submitted to legal experts for meticulous manual review, refinement, and auditing, ensuring a deterministic and legally sound basis for evaluating complex temporal reasoning.

\section{Details of Evaluation Prompts}
\label{sec:evaluation prompts}

\begin{table}[htbp]
    \centering
    \small
    \renewcommand{\arraystretch}{1.2}
    \setlength{\tabcolsep}{3pt}
    \begin{tabularx}{\columnwidth}{X}
        \toprule
        \cellcolor{promptgray} \textbf{\textsc{System:}} You are a legal Q\&A assistant. Please answer the effective date of the corresponding laws and regulations based on the question. (Note: If the legal document in the question does not specify a revision date or version, it indicates the original version.) The answer format is YYYY-MM-DD, without any extra explanation. \\
        \midrule
        \cellcolor{promptgray} \textbf{\textsc{Input:}} When did the ``Arbitration Law of the People's Republic of China (2009 Amendment)'' come into effect? \\
        \midrule
        \cellcolor{promptgray}\textbf{\textsc{Answer:}} 2009-08-27 \\
        \bottomrule
    \end{tabularx}
    \caption{The instruction and an example of Task 1-1: Effective Date Inquiry.}
    \label{tab:task_1_1}
\end{table}

\begin{table}[htbp]
    \centering
    \small
    \renewcommand{\arraystretch}{1.2}
    \begin{tabularx}{\columnwidth}{X}
        \toprule
        \cellcolor{promptgray} \textbf{\textsc{System:}} You are a legal Q\&A assistant. Based on the question, directly output the full name of the valid legal version without any other text. (Note: If the valid version is the original version of the law, you do not need to specify its revision date or version; just output the document name directly.) \\
        \midrule
        \cellcolor{promptgray} \textbf{\textsc{Input:}} On 1998-03-02, which version of the ``Arbitration Law of the People's Republic of China'' was active and valid? \\
        \midrule
        \cellcolor{promptgray}\textbf{\textsc{Answer:}} Arbitration Law of the People's Republic of China \\
        \bottomrule
    \end{tabularx}
    \caption{The instruction and an example of Task 1-2: Effective Version Identification.}
    \label{tab:task_1_2}
\end{table}

\begin{table}[htbp]
    \centering
    \small
    \renewcommand{\arraystretch}{1.2}
    \begin{tabularx}{\columnwidth}{X}
        \toprule
        \cellcolor{promptgray} \textbf{\textsc{System:}} You are a legal Q\&A assistant. For the given legal document, answer whether it was active and valid on the specified date. (Note: If the legal document in the question does not specify a revision date or version, it indicates the original version.) If it was valid, answer ``Yes''; if not, answer ``No''. Do not provide any additional explanation. \\
        \midrule
        \cellcolor{promptgray} \textbf{\textsc{Input:}} On 1995-07-25, was the ``Arbitration Law of the People's Republic of China'' active and valid? \\
        \midrule
        \cellcolor{promptgray}\textbf{\textsc{Answer:}} No \\
        \bottomrule
    \end{tabularx}
    \caption{The instruction and an example of Task 1-3: Effective Status Verification.}
    \label{tab:task_1_3}
\end{table}

\label{sec:a2_reasoning_tasks}

\begin{table}[htbp]
    \centering
    \small
    \renewcommand{\arraystretch}{1.2}
    \begin{tabularx}{\columnwidth}{X}
        \toprule
        \cellcolor{promptgray} \textbf{\textsc{System:}} You are a professional legal logical analysis assistant. Please accurately determine the chronological relationship between the two legal events A and B based on the provided ``case facts'' record. 
        The criteria are as follows:
        A. Sequential independent relationship: Event A strictly occurs before Event B, with a gap of one day or more (i.e., at least one day between the end of A and the start of B).
        B. Boundary connection relationship: Immediately after one event ends, another starts on the same day (e.g., B starts on the day A ends, or A starts on the day B ends).
        C. Internal inclusion relationship: The time span of a longer event strictly covers another shorter event.
        D. Sequential independent relationship: Event B strictly occurs before Event A, with a gap of one day or more (i.e., at least one day between the end of B and the start of A).
        Note: You are only allowed to output a single uppercase letter (A, B, C, or D). Do not output any explanation, inference, or multiple letters.
        Example reply: A \\
        \midrule
        \cellcolor{promptgray} \textbf{\textsc{Input:}} [Case Facts] 
        After trial, the court finds the following facts: On November 10, 2014, the plaintiff and the defendant signed a ``Labor Service Agreement,'' under which the defendant hired the plaintiff to work in a logistics service position. The term was two years, from November 10, 2014, to November 9, 2016. ...
        
        [Question]
        Based on the case record, please determine the chronological relationship between the following two items:
        Event A: Under the defendant's arrangement, the plaintiff continuously worked as a caregiver at the Third Affiliated Hospital of Zhejiang Chinese Medical University (Zhongshan Hospital).
        Event B: The plaintiff experienced back discomfort, underwent treatment in the massage department of the Third Affiliated Hospital of Zhejiang Chinese Medical University, and incurred medical expenses of 179 yuan.
        Please select the correct option:
        
        [Options]
        A. Sequential independent relationship (Event B occurs only after Event A ends)
        B. Boundary connection relationship
        C. Internal inclusion relationship
        D. Sequential independent relationship (Event A occurs only after Event B ends)
        
        Please directly output the unique uppercase letter of the answer. It is strictly forbidden to output any reasoning process or extra letters. \\
        \midrule
        \cellcolor{promptgray}\textbf{\textsc{Answer:}} C \\
        \bottomrule
    \end{tabularx}
    \caption{The instruction and an example of Task 2-1:Event  Relationship Discerning.}
    \label{tab:task_2_1}
\end{table}

\begin{table}[htbp]
    \centering
    
    \footnotesize 
    \renewcommand{\arraystretch}{1.2}
    \setlength{\tabcolsep}{3pt} 
    \begin{tabularx}{\columnwidth}{X}
        \toprule
        \cellcolor{promptgray} \textbf{System:} You are a legal case analysis assistant. Please calculate the absolute time difference between two events based on the given case facts, and select the correct or closest option. Only reply with the letter of the correct option, do not output extra explanation or text. \\
        \midrule
        \cellcolor{promptgray} \textbf{Input:} [Case Facts] Weng Shengrong was a excavator driver hired by the plaintiff. On June 11, 2015, Weng Shengrong was electrocuted while washing an excavator and died. On June 13, 2015, the plaintiff and the defendants signed a ``Compensation Agreement,'' agreeing that the plaintiff would compensate 850,000 yuan... \\
        \cellcolor{promptgray} [Question] How many days in total on the calendar elapsed from the day when ``Weng Shengrong was electrocuted and died while washing an excavator, and died after rescue efforts failed'' occurred to the day when ``the signed compensation agreement proved that the 600,000 yuan insurance payment should belong to the defendant'' occurred? \\
        \cellcolor{promptgray} [Options] A: 19 days \quad B: 26 days \quad C: 9 days \quad D: 12 days \\
        \midrule
        \cellcolor{promptgray} \textbf{Answer:} A \\
        \bottomrule
    \end{tabularx}
    \caption{The instruction and an example of Task 2-2: Temporal Distance Calculation.}
    \label{tab:task_2_2}
\end{table}

\begin{table}[htbp]
    \centering
    \small
    \renewcommand{\arraystretch}{1.2}
    \begin{tabularx}{\columnwidth}{X}
        \toprule
        \cellcolor{promptgray} \textbf{\textsc{System:}} You are a professional legal analysis assistant. Your task is to correctly sort the given events in chronological order based on the provided case facts record. Please read the case and question carefully. Note: You can only reply with a single uppercase letter of the correct option (e.g., A, B, C, or D). It is strictly forbidden to output any other punctuation, explanation, or extra characters. \\
        \midrule
        \cellcolor{promptgray} \textbf{\textsc{Input:}} [Case Facts] 
        The plaintiff Zheng Yanli claimed that: on May 12, 2017, at approximately 15:00, Li Xiaobao was driving a Toyota electric tricycle along the non-motorized lane... and collided with Zheng Yanli who was driving a Superqi electric two-wheeled bicycle, causing a traffic accident... Li Xiaobao was an employee of Zhenxing Wholesale Department...
        
        [Question]
        Based on the case facts record, please sort the following four events chronologically:
        ① ``Zheng Yanli's father Chen Yonggui died''
        ② ``Li Xiaobao signed a new employee entry agreement with Zhenxing Wholesale Department''
        ③ ``The judicial appraisal institution evaluated Zheng Yanli's disability level and nursing needs''
        ④ ``Li Xiaobao drove an electric tricycle and collided with Zheng Yanli driving an electric two-wheeled bicycle, causing a traffic accident''
        
        [Options]
        A: ③-④-②-①
        B: ①-②-④-③
        C: ②-①-③-④
        D: ②-④-③-① \\
        \midrule
            \cellcolor{promptgray}\textbf{\textsc{Answer:}} B \\
        \bottomrule
    \end{tabularx}
    \caption{The instruction and an example of Task 2-3: Global Event Ordering.}
    \label{tab:task_2_3}
\end{table}

\begin{table}[htbp]
    \centering
    \small
    \renewcommand{\arraystretch}{1.2}
    \begin{tabularx}{\columnwidth}{X}
        \toprule
        \cellcolor{promptgray} \textbf{\textsc{System:}} You are a professional legal analysis assistant. Your task is to select the unique correct event that meets the temporal requirements of the question from the given options based on the provided case facts record. Note: You can only reply with a single uppercase letter of the correct option (e.g., A, B, C, or D). It is strictly forbidden to output any other punctuation, explanation, or extra characters. \\
        \midrule
        \cellcolor{promptgray} \textbf{\textsc{Input:}} [Case Facts] 
        On March 6, 2017, the plaintiff Mei Xiuhua was injured by an electric arc shock from a high-voltage line while drying clothes on the balcony of her house. Mei Xiuhua was immediately hospitalized at Wuhan Third Hospital from March 6 to May 17, 2017. Later, from August 11, 2017 to August 21, 2017...
        
        [Question]
        According to the case facts, after ``Mei Xiuhua experienced an electric shock accident while using a mobile metal clothes hanger to dry clothes on the third-floor curved balcony of her house'', which of the following options represents the first (closest/earliest) subsequent event to occur?
        
        [Options]
        A. Mei Xiuhua was hospitalized at the Xinzhou District People's Hospital of Wuhan City due to shingles
        B. According to the case record, no subsequent event matching this condition was mentioned
        C. Wuhan Futian Aimin Judicial Appraisal Center appraised Mei Xiuhua's disability and evaluated her subsequent treatment costs
        D. Mei Xiuhua was hospitalized at the Wuhan Third Hospital due to residual wound surfaces \\
        \midrule
          \cellcolor{promptgray}  \textbf{\textsc{Answer:}} A \\
        \bottomrule
    \end{tabularx}
    \caption{The instruction and an example of Task 2-4-a: Bounded Event Retrieval (SLC).}
    \label{tab:task_2_4}
\end{table}

\begin{table}[htbp]
    \centering
    \small
    \renewcommand{\arraystretch}{1.2}
    \begin{tabularx}{\columnwidth}{X}
        \toprule
        \cellcolor{promptgray} \textbf{\textsc{System:}} You are a professional legal analysis assistant. Your task is to select all correct events that meet the temporal requirements of the question from the given options based on the provided case facts record. Note: This is a multiple-choice question. You may need to reply with one or more uppercase letters (e.g., A, AB, ABC). It is strictly forbidden to output any other punctuation, explanation, or extra characters. \\
        \midrule
        \cellcolor{promptgray} \textbf{\textsc{Input:}} [Case Facts] 
        On August 25, 2016, at approximately 15:00, the victim Wu was struck and killed by a falling storefront advertising sign. The storefront was owned by Textile Company.
        On September 28, 2015, Textile Company signed a lease contract with the non-party Xu Moujia...
    
        [Question]
        Combining the case facts and event logic, after the event ``The non-party Xu Moujia signed the 'Storefront Subcontract Agreement' with the defendant children's clothing store operator Xu Muhua'' and before the event ``The defendant children's clothing store operator Xu Muhua signed the 'Storefront Subcontract Agreement' with the non-party Cao Moujia'', which of the following events occurred during this period? (Multiple choice)
        
        [Options]
        A. The children's clothing store operator Xu Muhua paid a deposit to the defendant Huatai Advertising
        B. The defendant knew about the existence of the main frame of the storefront advertising sign through the contract or actual inspection, and tacitly assumed the safety management responsibility that should be borne as a user
        C. The children's clothing store operator Xu Muhua paid the remaining balance to the defendant Huatai Advertising
        D. The children's clothing store operator Xu Muhua paid 50,000 yuan in compensation to the agent of the victim Wu's mother, Wu Chunjiao \\
        \midrule
           \cellcolor{promptgray} \textbf{\textsc{Answer:}} AC \\
        \bottomrule
    \end{tabularx}
    \caption{The instruction and an example of Task 2-4-b: Bounded Event Retrieval (MLC).}
    \label{tab:task_2_5}
\end{table}

\begin{table}[htbp]
    \centering
    \small
    \renewcommand{\arraystretch}{1.2}
    \begin{tabularx}{\columnwidth}{X}
        \toprule
        \cellcolor{promptgray} \textbf{\textsc{System:}} You are a legal analysis assistant. Based on the given case facts and rights information, determine whether the right's claim has passed the peremptory period (immutable period). Only answer ``Yes'', ``No'', or ``Cannot determine'' . Do not include any extra text explanations. \\
        \midrule
        \cellcolor{promptgray} \textbf{\textsc{Input:}} [Case Facts] 
        The original court ruled to rescind the ``Guangdong Province Commodity House Sale Contract'' signed on November 8, 2012 between Wang Yuan and Country Garden, ordering Country Garden to return the purchase price of 5,500,000 yuan and pay a liquidated damage of 825,000 yuan...
        
        [Right under Review]
        Contract Rescission Right (Right to Terminate Contract)
        
        [Question]
        Please combine the facts to determine whether the claim date of this right has passed the ``peremptory period'' (immutable period).
        
        Decision logic:
        1. Answer ``Yes'': The starting point is clear, and the right claim date is later than the statutory deadline.
        2. Answer ``No'': The right claim date is before the statutory deadline, or is still within the active duration.
        3. Answer ``Cannot determine'': All other situations where the above conclusions cannot be deduced, including but not limited to insufficient information, active disputes, or the prerequisite right not being generated. \\
        \midrule
           \cellcolor{promptgray}\textbf{\textsc{Answer:}} No \\
        \bottomrule
    \end{tabularx}
    \caption{The instruction and an example of Task 3-1: Peremptory Period Reasoning.}
    \label{tab:task_3_1}
\end{table}

\begin{table}[htbp]
    \centering
    \small
    \renewcommand{\arraystretch}{1.2}
    \begin{tabularx}{\columnwidth}{X}
        \toprule
        \cellcolor{promptgray} \textbf{\textsc{System:}} You are a legal case analysis assistant. Your task is to determine whether the plaintiff's lawsuit has exceeded the statutory limitation of action period based on the provided case information. Only answer ``Yes'', ``No'', or ``Cannot determine''. Do not include any extra text explanations. \\
        \midrule
        \cellcolor{promptgray} \textbf{\textsc{Input:}} [Case Facts] 
        The original judgment found that: Du Yong and Du Chuandong are uncle and nephew. On September 20, 2009 (lunar calendar August 2nd), when a brick cave-dwelling built by Du Chuandong was undergoing construction, he invited clan members and relatives to help...
        
        [Question]
        Please determine whether the case has exceeded the limitation of action based on the case facts and legal provisions. The decision logic is as follows:
        1. Answer ``Yes'': The limitation of action period has expired. When the plaintiff exercised the right of claim, it had already exceeded the limitation of action period calculated from the statutory starting point, and there were no statutory circumstances of interruption, suspension, or extension.
        2. Answer ``No'': The limitation of action period has not expired. The date when the plaintiff exercised the right of claim falls within the statutory period, or events causing the interruption or suspension of the limitation of action occurred within the period, resulting in the period being recalculated or suspended.
        3. Answer ``Cannot determine'': It is impossible to reach any of the above definitive conclusions due to insufficient evidence, significant disputes over interruption events, or lack of information due to special limitation rules. \\
        \midrule
          \cellcolor{promptgray}  \textbf{\textsc{Answer:}} No \\
        \bottomrule
    \end{tabularx}
    \caption{The instruction and an example of Task 3-2: Action Limitation Reasoning.}
    \label{tab:task_3_2}
\end{table}

\begin{table}[htbp]
    \centering
    \small
    \renewcommand{\arraystretch}{1.2}
    \begin{tabularx}{\columnwidth}{X}
        \toprule
        \cellcolor{promptgray} \textbf{\textsc{System:}} You are a legal analysis assistant. Please analyze the given case facts and rights information according to the following logic chain, and ultimately determine whether the claim has exceeded the peremptory period (immutable period):
        
        [Reasoning Steps]
        1. Identify the rights and events, accurately lock in relevant time points, and clarify the applicable laws and regulations. Identify the type of rights involved and the nature of the event, preliminarily lock in the key time points of the case (such as the contract signing date, the date of the act, etc.), and determine the applicable version of the law according to the principle of non-retroactivity and relevant judicial interpretations.
        2. Determine the nature of the rights and clarify the length of the corresponding peremptory period. Based on the determined law or agreement, clarify the specific length of the peremptory period (such as ninety days, one year, five years, etc.). If the law clearly states that the period can be separately agreed upon, examine whether the parties have made a legal and valid agreement. If there is an agreement, it shall apply with priority; otherwise, the statutory period applies.
        3. Accurately locate the starting point of the peremptory period. Combined with the selected legal provisions or the parties' legal and valid agreement, identify and lock in the starting event of the peremptory period (e.g., from the date the party knew or should have known of the cause for rescission, from the date the legal act occurred, etc.).
        4. Compare the starting point with the right claim date to determine whether the period has elapsed. Calculate the time span from the determined starting point to the date the party actually claimed the right, and judge whether it exceeds the period length, thereby concluding whether the peremptory period has passed.
        
        [Output Format]
        First, please list the detailed analysis process of the above four steps. Write the analysis process after [Analysis Process].
        Finally, you must only answer ``Yes'', ``No'', or ``Cannot determine'' after [Answer], without any other explanations, notes, or extra words.
        Example: [Analysis Process] Step 1... Step 2... Step 3... Step 4... In summary, the right has expired. [Answer] Yes \\
        \midrule
        \cellcolor{promptgray} \textbf{\textsc{Input:}} [Case Facts]
        The original court ruled to rescind the ``Guangdong Province Commodity House Sale Contract'' signed on November 8, 2012 between Wang Yuan and Country Garden, ordering Country Garden to return the purchase price of 5,500,000 yuan and pay a liquidated damage of 825,000 yuan... 
        
        [Right under Review]
        Contract Rescission Right (Right to Terminate Contract)
        
        [Question]
        Please combine the facts to determine whether the claim date of this right has passed the ``peremptory period'' (immutable period). Please write the analysis process after [Analysis Process]. \\
        \midrule
        \cellcolor{promptgray}\textbf{\textsc{Answer:}} No \\
        \bottomrule
    \end{tabularx}
    \caption{The instruction and an example of Task 3-1: Peremptory Period Reasoning (Task-specific legal prompting).}
    \label{tab:task_3_1_ts}
\end{table}

\begin{table}[htbp]
    \centering
    \small
    \renewcommand{\arraystretch}{1.2}
    \begin{tabularx}{\columnwidth}{X}
        \toprule
        \cellcolor{promptgray} \textbf{\textsc{System:}} You are a legal analysis assistant. Your task is to analyze the case according to the following logic chain and determine whether the final claim has exceeded the limitation of action period.
        
        [Reasoning Steps]
        1. Determine the applicable law and transitional rules. Sort out key time points of the case (such as when the right was infringed, when the rights holder knew or should have known of the damage and the obligor, and when the performance period expired) to determine the applicable version of the law (such as the Civil Code, the original General Principles of the Civil Law, etc.).
        2. Identify the nature of the claim and determine the limitation period. Clarify the specific type of the involved claim. Based on the law determined in the previous step, determine whether the claim belongs to circumstances not subject to the limitation of action. If it does, stop the analysis and output ``Cannot determine''; if it is subject to the limitation of action, determine the specific period and specify the relevant legal provisions.
        3. Determine the starting point of the limitation of action. According to the applicable law, determine when the limitation period starts. Specify the exact starting date.
        4. Review whether there are any circumstances causing suspension or interruption of the limitation period according to the applicable law.
           - Interruption events: such as the rights holder requesting performance or the obligor agreeing to perform. Specify the timing of each interruption and list them in chronological order. After the interruption, the limitation period is recalculated from the end of the interruption event.
           - Suspension events: such as force majeure occurring within the last six months of the limitation period. If suspension exists, calculate the expiration date of the limitation period after the obstacle is removed.
        5. Calculate synthetically and determine whether the limitation of action has expired. Combining the starting point, interruption, and suspension circumstances, calculate the final expiration date, compare it with the date the rights holder claimed the right, and give a clear final determination.
        
        [Output Format]
        First, please list the detailed analysis process of the above five steps. Write the analysis process between [Analysis Process] and [Answer].
        Then, directly provide the final answer after [Answer], which must only be one of ``Yes'', ``No'', or ``Cannot determine'', without any other text.
        Example: [Analysis Process] Step 1... Step 2... Step 3... Step 4... Step 5... In summary, the limitation of action has expired. [Answer] Yes \\
        \midrule
        \cellcolor{promptgray} \textbf{\textsc{Input:}} [Case Facts] 
        The original judgment found that: Du Yong and Du Chuandong are uncle and nephew. On September 20, 2009 (lunar calendar August 2nd), when a brick cave-dwelling built by Du Chuandong was undergoing construction, he invited clan members and relatives to help...
        
        [Question]
        Please determine whether the case has exceeded the limitation of action based on the case facts and legal provisions. Please strictly follow the above five steps for analysis. \\
        \midrule
          \cellcolor{promptgray}  \textbf{\textsc{Answer:}} No \\
        \bottomrule
    \end{tabularx}
    \caption{The instruction and an example of Task 3-2: Action Limitation Reasoning (Task-specific legal prompt).}
    \label{tab:task_3_2_ts}
\end{table}

\end{document}